*Bogdan Czejdo, Wiktor B. Daszczuk, Waldemar Grabski, Sambit Bhattacharya*

# Cooperation of Multiple Autonomous Robots and Analysis of Their Swarm Behavior

*In this paper, we extended previous studies of cooperating autonomous robots to include situations when environmental changes and changes in the number of robots in the swarm can affect the efficiency to execute tasks assigned to the swarm of robots. We have presented a novel approach based on partition of the robot behavior. The sub-diagrams describing sub-routs allowed us to model advanced interactions between autonomous robots using limited number of state combinations avoiding combinatorial explosion of reachability. We identified the systems for which we can ensure the correctness of robots interactions. New techniques were presented to verify and analyze combined robots' behavior. The partitioned diagrams allowed us to model advanced interactions between autonomous robots and detect irregularities such as deadlocks, lack of termination etc. The techniques were presented to verify and analyze combined robots' behavior using model checking approach. The described system, Dedan verifier, is still under development. In the near future, timed and probabilistic verification are planned.*

### Introduction

The growing scope of applications of swarms of autonomous mobile devices (robots) is related with their natural ability to respond properly to malfunctions/collisions of individual robots and to environmental changes. The development of such systems of cooperating autonomous robots should have the high priority since they can be applied in many areas such as military reconnaissance, surveillance and guard systems, etc. The swarm robotics research has several challenges. The cooperation between several autonomous robots can be analyzed for a limited number of robots but scaling up more the solutions is an area that require new techniques. In this paper we concentrate on resilient cooperation of swarms of Robots based on partitioning behavior algorithms. We included techniques for automatic partial and total deadlock detection, and automatic checking inevitability of distributed termination, again partial and total. In the paper we attempt to answer several research questions.

a) How state diagrams can be used to efficiently describe the large number of robots?
b) How much we can scale up our solutions to guarantee the proper cooperation for the swarms of robots in terms of distributed termination and deadlock avoidance.
c) How much we can scale up our solutions to guarantee the proper cooperation for the robot swarms in terms of Mission objectives e.g. coverage of the area, frequency of checking each protected place etc.
d) How much we can scale up our solutions to guarantee the proper cooperation for the robot swarms, when we respond adequately to failures/collisions of individual robots and environmental changes.
e) How feasible is real-time verification of correctness of cooperation of autonomous moving platforms?

In this paper we concentrate on the problems of Autonomous Robots navigation in an indoor environment. We assume that the ROBOTs not only respond directly to the environment but also to actions of other robots. State diagrams have been previously used to describe the robots behavior [1]. Typically, the appropriate software is developed manually based on such models. To accelerate the development process we have created a new tool for the design of robots behavior and verification (Dedan [2]). Such a tool can be very useful for the rapid modification of robots reactive behavior since it allows the developers to incrementally modify the design. The tool functionality allow to automatically generate robots behavior in response to changing requirements.

In this paper we describe the techniques to analyze traditional state diagrams, to modify them to be more appropriate for swarm robotics and for integration of large number such state diagrams. The state diagrams allowed us to model advanced interactions between swarm of autonomous robots and can ensure the correctness of robots interactions. When rapid modifications of robots behavior are required, the rapid checking of robots interactions is crucial. The model checking method for state diagrams can identify problems such as deadlocks or live-locks and therefore offer the robots designer a set of ready-to-use algorithms and techniques for the analysis of complete swarm based system properties.

Deadlock freeness is checked by a CTL temporal formula **AG EX true** (for any state a next state exists) [3] [4] [5] [6]. However, partial deadlock cannot be so easily identified, therefore numerous methods for automatic detection of deadlocks in systems with a specific shape are proposed [7] [8] [9] [10] [11] [12].

In the analysis of distributed systems, two kinds of processes discontinuation are observed: undesired lack of progress (deadlock), which is an error, and expected stopping called process termination. Deadlock detection and termination detection methods must distinguish the two kinds of discontinuation [13], or simply prohibit one of them. However, total termination seems to be analogous situation: no future exists. In cyclic system, where termination is not expected, the above formula **AG EX true** identifies a deadlock. This is the reason that many deadlock detection techniques are addressed to endlessly looping systems only [9] [14] (discontinuation is a deadlock).

In terminating systems total deadlock should be distinguished from total termination. Various distributed termination detection techniques evolved [15] [16] [17] [18]. The methods are based on observation of some features of distributed processes or control over message traffic. Sometimes special elements of distributed processes are defined for termination detection.

Just as in a case of deadlock detection, dynamic (runtime) methods of termination detection require some instrumentation of a system. It is typically sending messages reporting the states of individual processes, and a mechanism of combining them into a global decision on distributed termination [13] [19] [20]. There are methods differing in instrumentation, dealing with failed processes or link failures, acceptance of temporary network partitioning [21] [22] [23] [24]. Static termination detection methods are based on observation of terminal states of individual processes. Model checking techniques are suitable for this purpose, using either model-specific formulas [25] or universal ones [26]. A construction of Counting Agent [27] may be applied both dynamically and statically. Transition invariants [28] allow to check if every execution starting in an initial state is finite.

Our approach is based on distinguishing terminating actions in a specification of distributed systems. Inevitability of termination is checked by a formula **AF** $(\varphi_1 \wedge \varphi_2 \wedge \ldots \wedge \varphi_n)$, where $\varphi_i$ denotes reaching of terminating action by $i^{th}$ robot. The sentence reads "the conjunction of $\varphi_1 \wedge \varphi_2 \wedge \ldots \wedge \varphi_n$ is inevitable", as **AF** $\varphi$ concerns eventual fulfilment of $\varphi$ on every execution path. Therefore, a partial termination may be tested easily by specifying of a subset of robots in the formula under **AF**. Even a single $i^{th}$ robot can be tested for its termination by **AF** $\varphi_i$.

Consider a topography presented in Fig. 1. It shows a couple of chambers with doors between them. Prefix A denotes side chambers while prefix Q denotes central chambers that are possibly in conflict during robots operation. The names of chambers are taken from cardinal directions. We assume that one robot may be present inside a typical chamber QNW, QNE, QSW and QSE at any time. There are also side chambers AW, AN, AE and AS that allow multiple robots co-locate. We also assume that each central chamber QNW, QNE, QSW and QSE has opening to two other central chambers and doors/opening for 2 side chambers.

## 2. Utilities for Rapid Generation of State Diagrams for Swarm Robot Navigation

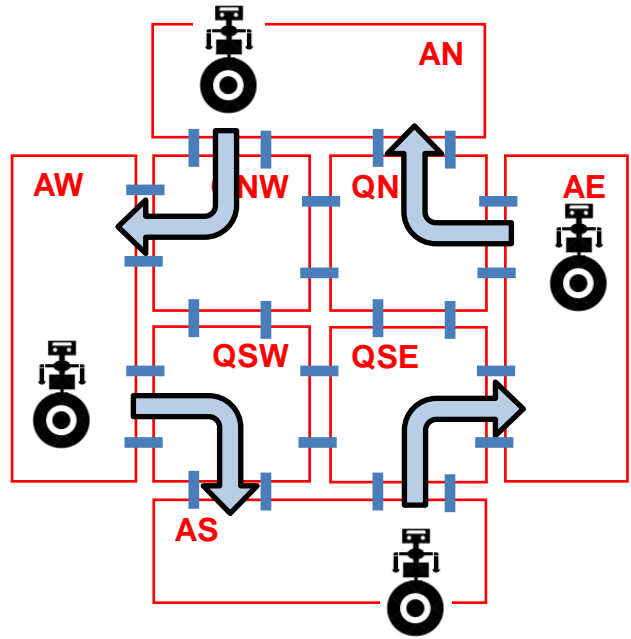

**Fig. 2** A Multiple-Robot Behavior

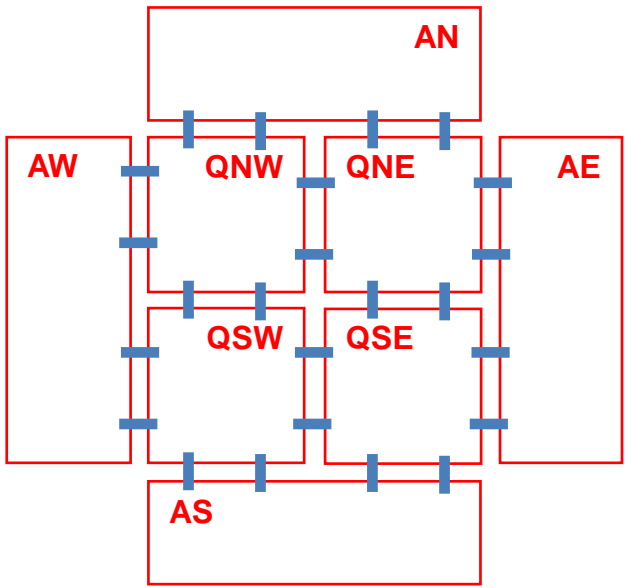

**Fig. 1.** An Environmental Resource Graph with 4 side chambers and 8 central chambers – resources that are in conflict

### 1. Environmental Resources

From the point of view of the robot programmer it is important to know the type of an environment the robots will move through. In general, an environment can be known or unknown. In this paper we will concentrate on describing robots behavior in a known environment. The known environment is typically described by a map identifying all parts of the building that we will refer as *chambers* and openings between chambers that we will refer simply as *doors*. One of the simple representations of a building structure can be a graph showing all accessible places in the form of nodes and the ways to get to these places in the form of graph paths designating order of accessing chambers.

Any topological map in the form of a graph can be also interpreted as a graph of environmental resources. It means that each node of the graph can be also interpreted as a resource and when the robot position is associated with this node we can claim that the robot acquired the resource. When the robot leaves the node we say that it releases the resource. The link between two nodes can be also interpreted as resource that can be acquired and released. Such an interpretation of a topological graph allows us to apply known resource allocation algorithms for the description of several robots behavior and extend it later to swarm of robots.

The deterministic state diagrams are well described in literature [29][30]. Generally, the deterministic state diagram, in addition to states, has transitions consisting of triggers that cause the transition of the robots from one state to another, and actions, that are invoked during a transition. Triggers are expressed by Boolean conditions evaluated continuously to respond to changes in the environment.

To specify state diagrams we use the notation based on Universal Modeling Language (UML) [31] where a state is indicated by a box and a transition is indicated by an arrow with a label. The first part of the transition label (before the slash) specifies the trigger and the part after the slash specifies the action (or message) to be invoked during the transition [31]. The syntax of probabilistic specifications is described in the literature [32] as an additional third component specifying the probability of the entire transition.

State diagrams that are explicitly location dependent can be convenient to specify robots behavior for several reasons. Firstly, the diagram can be constructed by relatively simple transformation of environmental resource diagram. Secondly, probabilistic components can be added relatively easily. Thirdly, the behavior of cooperating robots can be described by concurrent state diagrams and all well-established techniques for concurrent program analysis can be used i.e. deadlock detection or deadlock avoidance algorithms. The analysis of concurrency can be done automatically and the robot program can be directly generated from state diagram model.

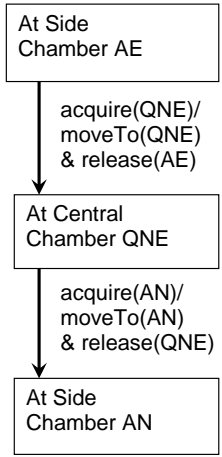

**Fig. 3.** A single robot behavior described by State Diagram

Based on environmental graph and corresponding environmental triggers we can rapidly specify various location-dependent state diagrams. Using autonomous robots, we may plan their behavior, specifying general rules they follow. A rule may be given as a target chamber for a robot, in the figure a robot starting from AS may be directed to AW, AN, AE, of back to AS, visiting the central chambers on the route. A set of routes may be independent of each other, as in a system of four patrol robots taking their direction always right (Fig. 2).

More precisely, the link between two nodes, e.g., AN and QNE can be interpreted as follows: if the robot is assigned a resource AN it should first acquire resource QNE before releasing resource AN. The state diagram shown in Fig. 2 specifies the ROBOT1 behavior A in some detail. Similarly the robots 2, 3 and 4 behavior can be described accordingly.

In order to formally specify such phrases as shown in Fig. 2, we need topological identification triggers, topological actions, and synchronization messages. Let us describe them in this order. Each of these topological constructs can be defined by a lower level diagrams.

Different topological places, i.e., different resources, would usually generate different values for the robot's sensors. The sensor signal processing algorithms i.e. algorithms describing a translation of robot sensor signals into a high level signals that can be used to directly identify the environment. We will assume that a lower level state diagram can describe such algorithm and we will refer to these signals to be used by a higher level diagrams as the environmental triggers.

To identify properly the solutions to our problems we will assume for further discussion the high level environmental trigger *acquire*() reflects the ability of robot's sensors and algorithms to recognize the chambers. Another high level environmental trigger *moveTo*() corresponds to the physical movement of vehicle from the actual location to the provided place, e.g., *moveTo(QNE)*.

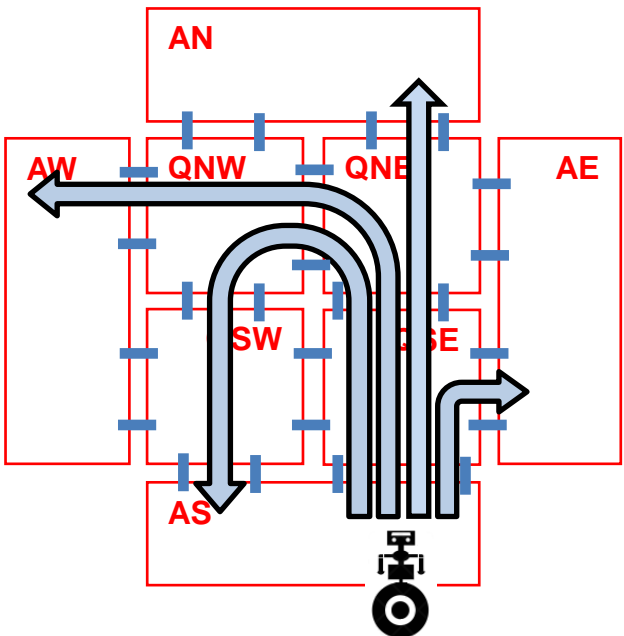

**Fig. 4.** A Multiple Robot Behavior described by a set State Diagrams created by *generate_many_behaviors*() utility function

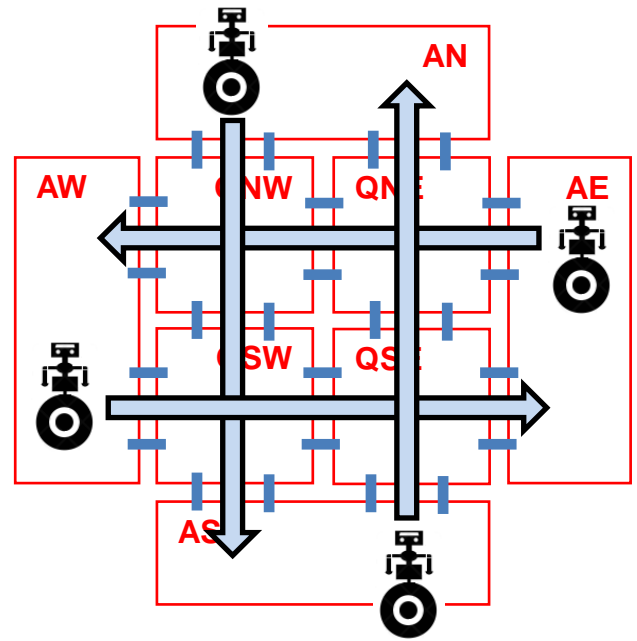

**Fig. 5.** A Multiple Robots Behavior that can result in a deadlock

For robots, let us consider behavior A describing a simple path for movement of ROBOT1 on the right in Fig. 2: start from the special chamber AE, then follow the door leading to QNE, then continue until entering AN and stop.

In order to model such behavior a state diagram model can be used. In general the multi-level model can be used [33], but in this paper for the simplicity of presentation, we assume two-level model. The upper level model is obtained by transforming the environmental graph i.e. converting non-directional to directional edges and providing the necessary triggers, actions and messages.

Creating efficiently such diagrams for multiple robots require some important utility functions. The first utility function in our project is to generate the identical behavior for different robots. We refer to this utility as *generate_identical_behavior*(behaviorN). It will generate a set of robots behavior with the same state diagrams. The second utility for our project is to generate the similar behavior that will have the same turns but will start from different chamber. We refer to this utility as *generate_similar_behavior*(behaviorP, positionXY). It will generate a set of robots behaviors that will have the same turns but will start from all possible side chambers. In addition to creation of complete behaviors, it is very important for our project to work on parts of diagrams and create a new set of diagrams based on partial behavior. We refer to these utility functions as *copy_part*() diagram and *append_identical_part*(...) and *append_similar_part*(...). Actually, the utility *generate_identical_*

*behavior*(behaviorN) and *generate_similar_behavior*(behaviorP, positionXY) can be special cases for *generate_identical_behavior* (behaviorN) and *generate_similar_behavior*(behaviorP, positionXY) when there is nothing to append. Fig. 3 shows a Multiple Robot Behavior described by a State Diagram created by *generate_similar_behavior*() utility function.

The presented utility functions allow us to generate rapidly all behaviors that can will need to be processed further as described in the next section. Let us mention yet another utility function *generate_all_behaviors*() and *generate_all_behaviors_from_ position*(...) that produce all possible behaviors but without cycles. The need for this utility will be clear in the next section. At this moment it is clear that we need to somehow restrict the behavior to avoid infinite state diagrams. Another option is to define *generate_ many*() to several behaviors with some restrictions. We are working on graphical interface for such utility functions to accelerate robot behavior specification even further. Fig. 4 shows an illustration of *generate_many_behaviors*() utility function where alternative behavior for the robot are be generated. It is obvious that such alternative behaviors are selected during run-time based on availability of chambers. There can be, however, important discussion about details of choice when several options are available e.g. based on some predefined order or even based on probabilities resulting in creation of probabilistic diagrams.

Yet, the routes may interfere as depicted in Fig. 5. An obvious total deadlock occurs if central chambers QNW, QNE, QSW and QSE are occupied.

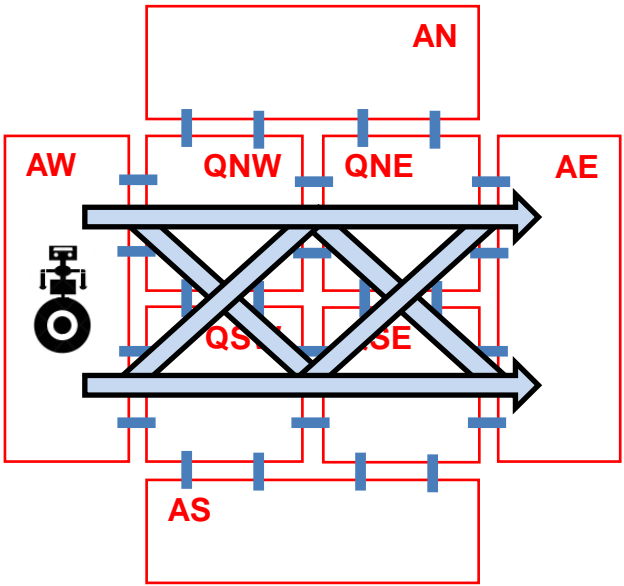

**Fig. 6.** A Multiple Robots Behavior that is free from deadlock

### 3. Cooperation between Multiple autonomous robots

In this section we discuss the cooperation between autonomous robots. There are many theoretical and practical solutions that can be taken into consideration.

The model checking provides a most general methodology [6] [34] [35] [36] [37] that can be used not only for deadlock avoidance or detection but also for detection and verification of wide variety of process cooperation characteristics. Typically the model checking is based on finite-state methods [35] that can be applied directly to our state diagrams. Therefore, it can be of important practical use for verifying some robot behaviors. Unfortunately, for the described asynchronous and probabilistic models the traditional model check-ing method cannot offer a set of ready-to-use algorithms and techniques for the analysis of properties of a multiple robots system.

Deadlock freeness can be easily proved using existing temporal verifiers like Spin [37], NuSMV [38] or Uppaal [39]. However, for efficient robot cooperation we need more subtle features. For example, partial deadlock is dangerous, because in such situation some robots continue their work but some of them are stuck and cannot do any progress. None of the mentioned popular model checkers find partial deadlocks automatically, a user must specify this feature themselves on a basis of features of a verified system.

Consider a topography described in previous sections and presented in Fig. 1. A set of routes may be independent of each other, as in a system of four patrol robots taking their direction always right (Fig. 3). Yet, the routes may interfere as depicted in Fig. 5. An obvious total deadlock occurs if chambers QNW, QNE, QSW and QSE are occupied. Such a deadlock may be avoided using routes that prevent a deadlock, for example allowing two robots starting from AW and AN, shown in Fig. 6, only some selected transitions. The other robots starting from AS and AE preserve full freedom of choice, as presented in Fig. 7.

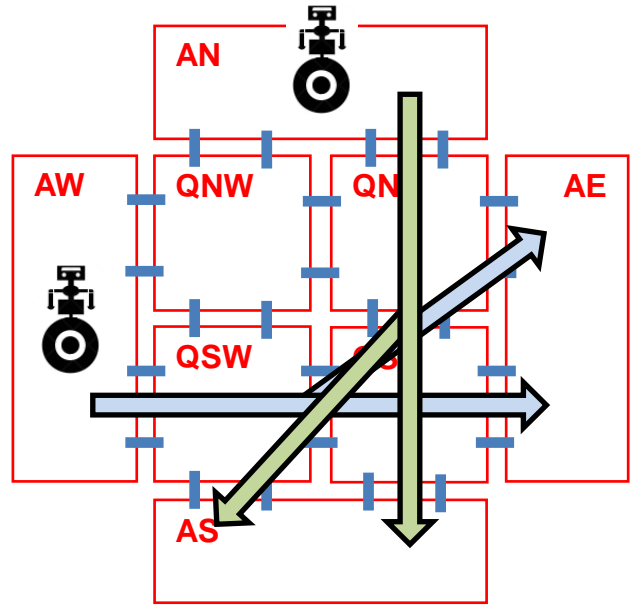

**Fig. 7.** Possible routes between side chambers AW and AE

A partial deadlock may occur if some robots block each other, as shown in Fig. 8. We need a tool for automated partial deadlock location. This may be performed for example in our IMDS formalism, in which a distributed system is specified as a set of nodes and a set of agents. We assume that every chamber is a node which controls its own occupancy, i.e., it may be in empty or occupied state. A robot may interrogate a node for a chamber occupancy state. This is achieved by issuing a message to a target chamber controller. If a positive answer is achieved (the chamber is empty), the robot sends a second message taking the target chamber then it frees the present chamber. In IMDS formalism, we can identify partial deadlocks in node/agent specification automatically [40] and reject conflicting routes of the robots.

However, total deadlock freeness and partial deadlock freeness are not enough for successful operation of multi-robot system. Consider Fig. 9: there are five robots, four of them are following a cycling patrol route AW, QSW, QSE, AE, QNE, AN, QNW, AW. If we add a fifth robot trying to move from AS to QSE, a chain of robots travelling through the chambers AW, QSW, QSE, AE may block it. No deadlock occurs, but the fifth robot (starting from AS) is stuck in fact.

Such a behavior may be identified using distributed termination feature. For this reason, we do not prepare cycling routes, instead we cut every route to a smaller sub-routes. These smaller sub-routes are chained to form a general behavior. On the end of every sub-route, a special terminating action is distinguished. Using such actions, the verifier may check if common distributed termination of a set of agents is achieved. The verification is performed using IMDS formalism, in which temporal formulas for total deadlock, partial deadlock and distributed termination are defined [40].

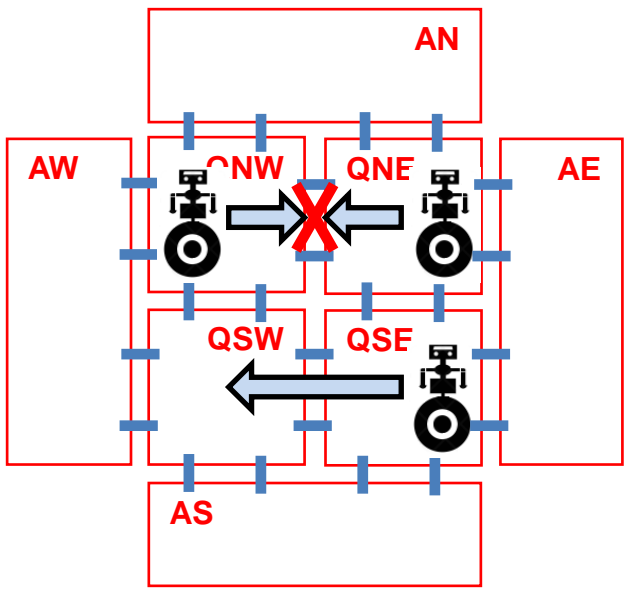

**Fig. 8.** Partial deadlock of two robots in central chambers QNW and QNE

The presented methodology assigns a fixed route to every robot. Conversely, we may allow the robots to take alternative path, which leaves much freedom of choice to autonomous robots. To avoid deadlocks, we must prohibit some conflicting sub-routes from a robot behavior plan. Consider a system in Fig. 7: a robot starting from AW may follow a route consisting in AW, QNW, QNE, AE or alternatively AW, QSW, QSE, AE. Also, from QNW it may diverge to QSW or reverse. The same concerns chambers QNE and QSE. For other robots the situation is symmetric. This leads to a deadlock for example when the robots choose the routes depicted in Fig. 5.

If we assign probabilities of taking individual transitions (for example for the agents behaving like in Fig. 7), Probability of a deadlock may be calculated, for example using Prism probabilistic model checker [41].

In the routes that are prone to deadlock, like in Fig. 9, a system may avoid this situation if proper timing is assumed for making transitions between chambers and maximum time spent in chambers. For verification of such systems, timed model checker (Uppaal or Prism) may be used.

If probabilistic or timed model checking is applied (or a combination of the two), and we combine it with IMDS specification, partial deadlocks can be found, communication deadlock can be distinguished from resource deadlock, and inevitability of termination can be checked.

### 4. Verification of swarm robot behavior in Dedan

The methodology presented in the previous section allows us to create sub-routes from a robot behavior plan. As a result, we can avoid deadlocks, by prohibiting some conflicting sub-routes from a robot behavior plan. There are two partition methods.

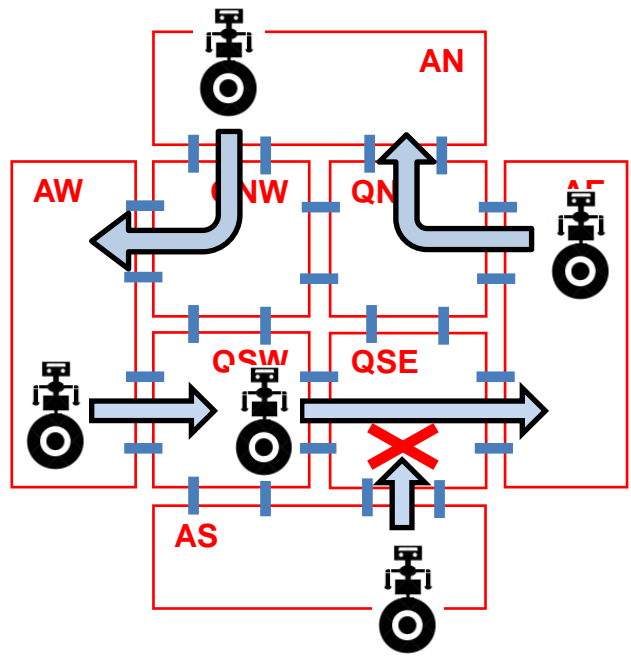

**Fig. 9.** A robot in AS which is stuck without a deadlock

The first partition assumes that the complete behavior can be partitioned into sub-routes with the beginning and the end at special chamber and with no cycles. This way we can deal with limited sub-behaviors and identify for all of them synchronization problems. This approach allow us to avoid exponential growth of as described in the Integrated Model of Distributed Systems (IMDS) formalism [42]. In this formalism a real distribution of elements may be expressed, since the actions of distributed elements are based on local states only. The Dedan verification environment, which uses IMDS specification, has been implemented to find deadlocks in cooperating distributed elements using model checking technique [43], Resource deadlocks and communication deadlocks, both total and partial, are searched automatically in Dedan.

Sometimes it is not sufficient to identify deadlocks in the system, therefore automatic termination checking is added to Dedan (as in Fig. 9, which presents non-terminating robot starting from AS, while no deadlock occurs).

Actually the first type of partition should be sufficient to cover most application since it can provide guaranteed coverage of all area and avoidance of synchronization problems. There need to be some adjustment for this method to allow for multiple execution of the same behavior for by many robots. We considered a novel technique for analyzing a sequence of robots with identical behavior following one another by representing a set of states occupied by the robot by a single superstrate to identify deadlocks.

The second type of partition requires breaking the cycle in the state related to the regular chamber. Such situation obviously provides many more sub-routs. Our approach was to add to each sub-route starting or finishing in the central chamber a path to a closest side chamber for analysis only to consider a worse case. Based on this approach the number of sub-routs to consider for analysis was not increased. The additional difficulty was to determine the numerical threshold for number of sub-routs that can be executed the same time, which is strongly related to number of robots in the swarm that can cooperate successfully the same time.

A distributed system is typically described in terms of servers exchanging messages. A process in such a system can be defined as a sequence of changes of a server states. The states of servers are internal to the processes, which communicate by the message exchanges (a client-server model [44]).

Such a principle is a basis of IMDS formalism, which defines a configuration of a distributed system as a set of current servers' states $p \in P$ and current messages $m \in M$, belonging to distributed computations, called agents. In our example, servers are chamber controllers while agents are robots travelling through them.

The dynamics of a distributed system is described as actions, having a pair ($m,p$) on input, which describes a state of a given server and a message of some agent pending at this server. The executed action changes a state $p$ of the servers to a new state $p'$. Likewise, a message $m$ disappears and a new message $m'$ is created in the context of the same agent. Therefore, we may treat an action as a relation $\Lambda \subset (M \times P) \times (M \times P)$.

An agent may terminate in a special action in which no message is generated. Terminating actions are in the product $(M \times P) \times (P)$.

Formally, a server state is a pair $p=$(server, state) and a message is a triple $m=$(agent, server, service). A process is a sequence of actions: in the same server (server process) or in the same agent (agent process). Formal definition of IMDS may be found in [40].

Let us analyze autonomous robots system when a robot tries to acquire a chamber. For example, if the agent ROBOT1 is in the side chamber AE, it tries to take central chamber QNE. To do it safely, first QNE is taken, and then AE is released. It is done by means of three messages:
1. The message 'try' is sent from the AE to QNE. This message may wait for acceptance for undefined period of time if QNE is occupied.
2. If at last the message 'try' is accepted in QNE is accepted (QNE is free at this time), the message 'ok' is sent back from QNE to AE. QNE changes its state from 'free' to 'reserved' – it cannot be taken by other robots.
3. Then, AE is released and QNE is finally taken by the ROBOT1, AE becomes 'free' and QNE becomes 'occupied'.

The deadlock may be observed from the servers' or from the agents' point of view. Dedan finds both kinds of deadlock (in communication and over resources) automatically. Likewise, inevitability of an agent's termination is automatically verified.

Below we present the Dedan code for a simplest case shown in Fig. 3. Two robots: ROBOT[1] and ROBOT[2] travel in opposite directions between two side chambers, having a central chamber in between. This obviously leads to a deadlock. Yet, we show a counterexample of termination checking for agent ROBOT[1]: it shows the failure of reaching the robot's target as a sequence diagram.

```
#DEFINE N 2

server:   SideCh(agents ROBOT[N];servers CentralCh),
//Side Chamber
services  {start,tryS[2],okS[2],takeS},
//S - going from Central Chamber
//try - test ok access, ok - accept, take - enter
states    {free,resS,occ,end},
//free - free, res - reserved, occ - occupied
actions{
<i=1..N> {ROBOT[i].SideCh.start, SideCh.occ} ->
         {ROBOT[i].CentralCh.tryC[i], SideCh.occ},
<i=1..N><j=1..2>{ROBOT[i].SideCh.okS[j], SideCh.occ} ->
         {ROBOT[i].CentralCh.takeC[j], SideCh.free},

<i=1..N><j=1..2>{ROBOT[i].SideCh.tryS[j], SideCh.free} ->
         {ROBOT[i].CentralCh.okC[j], SideCh.resS},

<i=1..N><j=1..2>{ROBOT[i].SideCh.tryS[j], SideCh.occ} ->
         {ROBOT[i].CentralCh.notC[j], SideCh.occ},
<i=1..N> {ROBOT[i].SideCh.takeS, SideCh.resS} ->
         {SideCh.end},
}

server:   CentralCh(agents ROBOT[N];servers SideCh[2]),
//Central Chamber
services  {tryC[2],okC[2],notC[2],takeC[2],switch[2]},
states    {free,resC[2],occ},
actions{
//going to Side Chamber
<i=1..N><j=1..2>{ROBOT[i].CentralCh.tryC[j], CentralCh.free} ->
         {ROBOT[i].SideCh[j].okS[j], CentralCh.resC[j]},
<i=1..N><j=1..2>{ROBOT[i].CentralCh.takeC[j], CentralCh.resC[j]}->
         {ROBOT[i].CentralCh.switch[3-j], CentralCh.occ},
<i=1..N><j=1..2>{ROBOT[i].CentralCh.switch[j], CentralCh.occ} ->
         {ROBOT[i].SideCh[j].tryS[j], CentralCh.occ},
<i=1..N><j=1..2>{ROBOT[i].CentralCh.okC[j], CentralCh.occ} ->
         {ROBOT[i].SideCh[j].takeS, CentralCh.free},
}

servers   SideCh[2],CentralCh;
agents    ROBOT[N];

init->    {
          <j=1..2>SideCh[j](ROBOT[1..N],CentralCh).occ,
             CentralCh(ROBOT[1..N],SideCh[1,2]).free,

          <j=1..2>ROBOT[j].SideCh[j].start,
          }.
```

A counterexample showing a lack of termination of ROBOT[1] is presented in Fig. 10.

**Conclusions**

In this paper, we extended previous studies of cooperating autonomous robots to indoor environments and include situations when environmental changes and changes in the number of robots in the swarm can improve or make worse the efficiency to execute tasks assigned to the swarm of robots. We have presented a novel approach using partition of the robot behavior. The sub-diagrams describing robots behavior allowed us to model advanced interactions between autonomous robots based on limited number of state combinations avoiding state explosion. We identify the systems for which we can ensure the correctness of robot interactions and the techniques were presented to verify and analyze combined robots' behaviors.

The Dedan verification environment is using model checking technique, for finding communication deadlocks and resource deadlocks, partial and total. Also, distributed termination is verified automatically. Moreover, the system may be automatically converted from the server view to the agent view, the state space of the system may be observed and simulated, and the system may be converted to Promela (Spin verifier input form [25] [37]) and Uppaal [39]. A graphical form of verified system representation is possible and graphical simulation over component servers/agents are supported. The described system is still under development. In near future, an own algorithm for non-exhaustive partial deadlock search will be included. A new concept of distributed automata is under development. More advanced forms of verification will be available, using timed automata ([45][46], to verify real-time dependencies), and probabilistic model checking [41]. One of the most advanced features will be automatic or semi-automatic behavior modification that will significantly improve the dynamic resilience of cooperating autonomous robots.

**Acknowledgment**

Bogdan Czejdo's research was partially supported by the Belk Foundation.

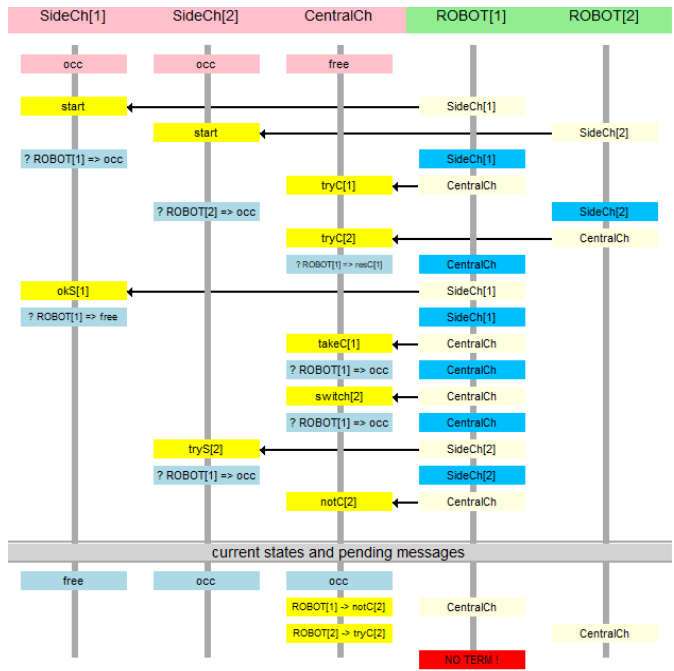

**Fig. 10**. Counterexample showing a lack of termination of ROBOT[1]

---

**Współpraca Roju Autonomicznych Robotów i Analiza Ich Zbiorowych Zachowań**

W artykule opisano kontynuację wcześniejszych badań dotyczących współpracy autonomicznych robotów wewnątrz budynku. Obejmują one obejmują sytuacje, w których zmiany środowiska i zmiana liczby robotów w roju mogą poprawić lub pogorszyć efektywność wykonywania zadań przypisanych do roju robotów. Zaprezentowaliśmy nowatorskie podejście z wykorzystaniem dzielenia zachowań robota na zachowania składowe. Pod-diagramy opisujące kładowe pod-marszruty pozwoliły nam modelować zaawansowane interakcje między autonomicznymi robotami w oparciu o ograniczoną liczbę kombinacji zachowań, unikając eksplozji kombinatorycznej przestrzeni osiągalności. Opisano systemy, dla których możemy zapewnić poprawność interakcji robotów i zaprezentowano techniki weryfikacji i analizy zachowań połączonych robotów. Diagramy podzielone na partycje pozwoliły nam modelować zaawansowane interakcje pomiędzy autonomicznymi robotami i wykrywać nieprawidłowości, takie jak zakleszczenia, brak terminacji itp. Przedstawiono techniki weryfikacji i analizy złożonych zachowań robotów za pomocą techniki weryfikacji modelowej. Opisany system weryfikacji, Dedan, jest wciąż rozwijany. W niedalekiej przyszłości planowana jest weryfikacja z czasem rzeczywistym i probabilistyczna.

**Słowa kluczowe:** autonomiczne roboty, weryfikacja zachowań, weryfikacja modelowa, Zintegrowany Model Systemów Rozproszonych, Integrated Model of Distributed Systems, zakleszczenie, terminacja.

**Authors:**

**Bogdan Czejdo**, PhD - Department of Mathematics and Computer Science, Fayetteville State University, Fayetteville, NC 28301, USA, bczejdo@uncfsu.edu

**Wiktor B. Daszczuk**, PhD – Warsaw University of Technology, Institute of Computer Science, Nowowiejska Str. 15/19, 00-665 Warsaw, Poland, wbd@ii.pw.edu.pl

**Waldemar Grabski**, MSc – Warsaw University of Technology, Institute of Computer Science, Nowowiejska Str. 15/19, 00-665 Warsaw, Poland, wgr@ii.pw.edu.pl

**Sambit Bhattacharya**, PhD - Department of Mathematics and Computer Science, Fayetteville State University, Fayetteville, NC 28301, USA, sbhattac@uncfsu.edu